\documentclass[conference]{IEEEtran}

\ifCLASSINFOpdf
\else
\fi
\usepackage{subcaption}
\usepackage{caption}
\usepackage{graphicx}
\usepackage[ruled]{algorithm2e}
\usepackage{mathtools}
\usepackage{epsfig}
\usepackage{lipsum}
\usepackage{float}
\usepackage{dblfloatfix}
\usepackage{color}
\usepackage{times}
\usepackage{amsmath}
\usepackage{amssymb}

\IEEEoverridecommandlockouts
\begin{document}
%
\title{Autonomously and Simultaneously Refining Deep Neural Network Parameters by a Bi-Generative Adversarial Network Aided Genetic Algorithm}

\author{Yantao~Lu,
        Burak Kakillioglu,
        Senem~Velipasalar \\
        Department of Electrical Engineering and Computer Science\\
        Syracuse University\\
        Syracuse NY, 13244-1240\\
        e-mail: ylu25@syr.edu, bkakilli@syr.edu, svelipas@syr.edu
        \thanks{This work has been funded in part by National Science Foundation (NSF) under CAREER grant CNS-1206291 and NSF Grant CNS-1302559.}
}


\maketitle

\begin{abstract}
The choice of parameters, and the design of the network architecture are important factors affecting the performance of deep neural networks. Genetic Algorithms (GA) have been used before to determine parameters of a network. Yet, GAs perform a finite search over a discrete set of pre-defined candidates, and cannot, in general, generate unseen configurations. In this paper, to move from exploration to exploitation, we propose a novel and systematic method that autonomously and simultaneously optimizes multiple parameters of any deep neural network by using a GA aided by a bi-generative adversarial network (Bi-GAN). The proposed Bi-GAN allows the autonomous exploitation and choice of the number of neurons, for fully-connected layers, and number of filters, for convolutional layers, from a large range of values. Our proposed Bi-GAN involves two generators, and two different models compete and improve each other progressively with a GAN-based strategy to optimize the networks during GA evolution. Our proposed approach can be used to autonomously refine the number of convolutional layers and dense layers, number and size of kernels, and the number of neurons for the dense layers; choose the type of the activation function; and decide whether to use dropout and batch normalization or not, to improve the accuracy of different deep neural network architectures. Without loss of generality, the proposed method has been tested with the ModelNet database, and compared with the 3D Shapenets and two GA-only methods. The results show that the presented approach can simultaneously and successfully optimize multiple neural network parameters, and achieve higher accuracy even with shallower networks.
\end{abstract}

%


%
\IEEEpeerreviewmaketitle

\section{Introduction}
Deep learning-based techniques have found widespread use in machine learning. Even before the convolutional approaches becoming popular, simple multi-layer perceptron neural networks (MLPNN) were widely used for classification tasks for several reasons. First, they are very easy to construct and run since each layer is represented and operated as a single matrix multiplication. Second, a neural network can become a complex non-linear mapping between input and output by the introduction of non-linear activation functions after each layer. Regardless of how large the input size (i.e. the number of features) or the output size (i.e. the number of classes) are, a neural network can discover relations between them when the network is sufficiently large; and enough samples, which cover the problem domain as much as possible, are provided during training.

Although a MLPNN is successful to form a complex non-linear relationship between the feature space and the output class space, it lacks the ability of discovering features by itself. Until recent years, the classical approach was to provide either the input data directly or some high-level descriptors, which are extracted by applying some algorithm on the input data, as the source of features. Using the raw input as features does not guarantee to yield any satisfactory mappings and the latter approach would require the investigation of multiple hand-crafted descriptor extraction algorithms for different applications. The introduction of convolutional layers in neural networks removed the necessity of having prior feature extractors, which are not easy to craft for different applications. Convolutional layers are designed to extract features directly from the input. Since they have been proven to be successful feature extractors and thanks to much faster computation of the operations of convolutional neural networks (CNNs) on specialized processors such as GPUs, the use of CNNs exploded recently. After Krizhevksy et al.~\cite{krizhevsky2012imagenet} achieved a significant increase in the classification accuracy on the ImageNet Large Scale Visual Recognition Challenge (ILSVRC)~\cite{deng2009imagenet} in 2012, many others followed their approach, by creating different architectures and applying them to numerous applications in many different domains.

It is well-known that the training of deep learning methods requires large amounts of data, and they usually perform better when training data size is increased. However, for some applications, it is not always possible to obtain more data when the dataset at hand is not large enough. In many cases, even though the raw data can be collected easily, the labeling or annotation of the data is difficult, expensive and time consuming. Successors of ~\cite{krizhevsky2012imagenet} yielded better accuracies with less number of parameters on the same benchmark with some architectural modifications using the same building blocks. This shows that the choice of parameters, and the design of the architecture are important factors affecting the performance. In fact, the design of a CNN model is very important to achieve better results, and many researchers have been working hard to find better CNN architectures \cite{simonyan2014very,zeiler2014visualizing,lin2013network,szegedy2015going,szegedy2016rethinking,he2016deep,ren2015faster} to achieve higher accuracy.

However, there has not been much work on developing an established and systematic way of building the structure of a neural network, and this task heavily depends on trial and error, empirical results, and the designer's experience. Considering that there are many design and parameter choices, such as the number of layers, number of neurons in each layer, number of filters at each layer, the type activation function, the choice of using drop out or not and so on, it is not possible to cover every possibility, and it is very hard to find the optimal structure. In fact, often times some common settings are used without even trying different ones. Moreover, the hyper-parameters in training phase also play important role on how well the model will perform. Likewise, these parameters are also tuned manually in an empirical way most of the time.

In this work, we focus on optimizing the network architecture and its different parameters for any neural network model. We propose a novel and systematic way, which employs a revised generative adversarial networks, referred to as Bi-GAN, together with a Genetic Algorithm (GA). The proposed method can autonomously refine the number of convolutional layers, the number and size of filters, number of dense layers, and number of neurons; decide whether to use batch normalization and max pooling; choose the type of the activation function; and decide whether to use dropout or not.

\subsection{Related Work}
There have been works focusing on optimizing neural network architectures. Most of the proposed approaches are based on the GAs~\cite{Deb2002}, or evolutionary algorithms, which are heuristic search algorithms. Benardos and Vosniakos~\cite{benardos2007optimizing} proposed a methodology for determining the best neural network architecture based on the use of a GA and a criterion that quantifies the performance and the complexity of a network. In their work, they focus on optimizing four architecture decisions, which are the number of layers, the number of neurons in each layer, the activation function in each layer, and the optimization function. Magnier and Haghighat~\cite{Laurent2010} presented an optimization methodology based on a combination of a neural network and a multi-objective evolutionary algorithm. The methodology was used for the optimization of thermal comfort and energy consumption in a residential house. Leung et al.~\cite{Leung2003} presented the tuning of the structure and parameters of a neural network using an improved GA. Ritchie et al.~\cite{ritchie2003optimizationof} proposed a method to automate neural network architecture design process for a given dataset by using genetic programming. Islam et al.~\cite{islam2014optimization} employed a genetic algorithm for finding the optimal number of neurons in the input and hidden layers. They apply their approach to power load prediction task and report better results than a manually designed neural network. Yet, their approach is used to optimize only the number of neurons for input and hidden layers, and optimization of other important design decisions such as the number of layers or activation function type are not discussed. Stanley and Miikkulainen~\cite{stanley2002evolving} presented the NEAT algorithm for optimizing neural networks by evolving topologies and weights of relatively small recurrent networks. In a recent work, Miikkulainen et al.~\cite{miikkulainen2017evolving} proposed CoDeepNEAT algorithm for optimizing deep learning architectures through evolution by extending existing neuroevolution methods to topology, components and hyperparameters.

The genetic algorithm-based optimization uses a given set of blueprints and models, i.e. it performs a finite search over a discrete set of candidates. Thus, GAs, in general, cannot generate unseen configurations, and they can only make a combination of preset parameters. GAs are good at searching better solutions from limited possibilities, such as type of layers and activation functions. However, it cannot search for a solution which is not defined before. In addition, the complexity of GAs increases significantly when the number of choices increases to large scale. Rylander~\cite{Rylander:2001} has shown that the generations needed for convergence increases exponentially with the node size.

Apart from the genetic algorithms, Bergstra and Bengio~\cite{bergstra2012random} have proposed random search for hyper-parameter optimization, and stated that randomly chosen trials are more efficient for hyperparameter optimization than trials on a grid. Yan and Zhang~\cite{jin2016neural} optimized architectures' width and height with growing running time budget through submodularity and suparmodularity.
\begin{figure*}[htb!]
\centering
\includegraphics[width=.55\textwidth]{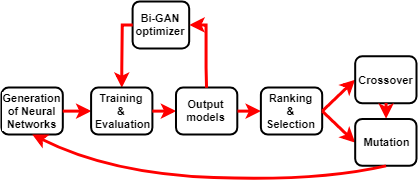}
\caption{Proposed Bi-GAN aided GA network for refining deep neural network parameters.}
\label{fig:GA-GAN}
\end{figure*}

Generative Adversarial Networks (GANs)~\cite{goodfellow2014generative} are one of the important milestones in deep learning research. In contrast to CNNs, which extract rich and dense representations of the source domain, and may eventually map source instances into some classes, GANs generate instances of the source domain from small noise. They employ deconvolution operators, or transposed convolutions, to generate N-D instances from 1-D noise. GAN's power comes from the competition with the discriminator, which decides whether the generated instance belongs to the source domain. Discriminator acts like the police who is trying to intercept counterfeit money, where in this case the generator is the counterfeiter. Generator and discriminator are trained together until discriminator cannot distinguish the generated instances from the instances in the source domain. GANs have been adapted in many applications~\cite{gatys2015neural,radford2015unsupervised,johnson2016perceptual,isola2017image,zhu2017unpaired}.

In this paper, to move from exploration to exploitation, we propose a novel and systematic method that autonomously and simultaneously optimizes multiple parameters of any deep neural network by using a GA aided by our proposed bi-generative adversarial network (Bi-GAN). In contrast to traditional GANs, our proposed Bi-GAN involves two generators, and two different models compete and improve each other progressively with a GAN-based strategy to optimize the networks during GA evolutions. The Bi-GAN allows the autonomous exploitation and choice of the number of filters and number of neurons from a large range of values. Our proposed approach can be used to autonomously refine the number of convolutional layers, the number and size of filters, number of dense layers, and number of neurons; decide whether to use batch normalization and max pooling; choose the type of the activation function; and decide whether to use dropout or not, to improve the accuracy of different deep neural network architectures.

For this work, without loss of generality, we tested the performance of our approach by using the ModelNet database, and compared it with the 3D Shapenets and two GA-only methods. The results show that the presented approach can simultaneously and successfully optimize multiple neural network parameters, and achieve increased accuracy even with shallower networks. The rest of this paper is organized as follows: The proposed method is described in Sec.~\ref{sec:proposed}. The experimental results are presented in Sec.~\ref{sec:exp}, and the paper is concluded in Sec.~\ref{sec:concl}.

\section{Proposed Method}\label{sec:proposed}
The overall structure of the proposed method is shown in Fig.~\ref{fig:GA-GAN}. It involves a GA aided by a Bi-GAN. While Bi-GAN is used to set/optimize the parameters from a large set covering a large range, GA is applied to make discrete decision from a small set of choices. The decisions the GA makes include the choice of the activation function; whether or not to use batch normalization, dropout and max pooling; number of convolutional layers and dense layers; and the kernel size of convolutional layers. Table \ref{table:GA parameters} shows the set of parameters, and the discrete set of values that they can take.
\begin{table}[h!]
{\small{
\caption{Parameter Choices}
\vspace{0.1cm}
\begin{tabular}{ll}
Activation function   & \{``relu", ``leaky relu", ``sigm.", ``tanh"\} \\
Batch norm.    & \{True, False\}                             \\
Dropout                & \{True, False\}                             \\
Max pooling            & \{True, False\}                             \\
Num. of conv layers  & \{1,2,3\}                                   \\
Num. of dense layers & \{1,2,3\} \\
Kernel size             &  \{3,5\}
\end{tabular}
\label{table:GA parameters}
}}
\end{table}

The parameter set of the network $i$ for the GA has the following form: \\
\begin{equation}
\begin{split}
p_i^{GA} = &[prm_i^{conv^1},prm_i^{conv^2}..., prm_i^{conv^C}, \\
&prm_i^{dense^1},prm_i^{dense^2}..., prm_i^{dense^D} ],\\
&i \in \{1,2,...,n_m\}
\end{split}
\label{eqn:GA_param1}
\end{equation}
%
\noindent where $C$ and $D$ are the \textit{maximum} number of possible convolutional layers and dense layers, respectively, $n_m$ is the number of network models in the population and
\begin{equation}
\begin{split}
prm_i^{conv^j}=&[ 1/0\text{ (conv. layer exists or not)}, \text{kernel size}, \\
& \text{activation func.}, 1/0\text{ (for batch norm.)}, \\
&1/0\text{ (for max pooling)}], \\
&i \in \{1,2,...,n_m\}\\
&j \in \{1,2,...,C\}
\end{split}
\label{eqn:GA_conv_prm}
\end{equation}
\begin{equation}
\begin{split}
prm_i^{dense^k}=&[1/0\text{ (dense layer exists or not)}, \\
&\text{activation func.}, 1/0 \text{ (for batch norm.)},\\
&1/0\text{(for dropout)}], \\
&i \in \{1,2,...,n_m\}\\
&k \in \{1,2,...,D\}.
\end{split}
\label{eqn:GA_dense_prm}
\end{equation}
For the discrete set of parameters given in Table~\ref{table:GA parameters}, the values of $C$ and $D$ are $3$.

First, $n_m$-many models are randomly initialized for the first network population $N^1$. This is done by choosing the values of convolutional and dense parameters, in (\ref{eqn:GA_conv_prm}) and (\ref{eqn:GA_dense_prm}), randomly, from the possible choices. Then, the number of filters for the convolutional layers, and the number of neurons for the fully connected layers are determined by the Bi-GAN as will be described in Section \ref{ssec:Bigan}.

Then, the $n_m$ models are trained, and evaluated to obtain their accuracy scores. Based on the accuracy scores, the GA is applied as detailed in Section \ref{ssec:GA}.


\subsection{Bi-GAN network} \label{ssec:Bigan}

\begin{figure}[htb!]
\centering
\includegraphics[width=.5\textwidth]{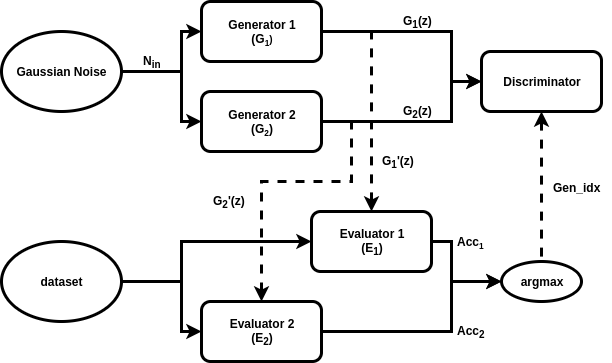}
\caption{Proposed Bi-GAN incorporating two generators and one discriminator.}
\label{fig:biGAN}
\vspace{-0.2cm}
\end{figure}

We propose a novel and modified generative adversarial network (GAN), referred to as Bi-GAN, to find the optimal network parameters, that have a large range of values. The proposed Bi-GAN network for refining different neural network parameters is shown in Fig.~\ref{fig:biGAN}. It is composed of a generative part, an evaluation part and a discriminator. In contrast to traditional GANs, there are two generators ($G_1$ and $G_2$), two evaluators ($E_1$ and $E_2$), and one discriminator ($D$). The input to the two generators is Gaussian noise $z \sim p_{noise}(z)$. On the other hand, the input to the evaluators is the training data $x \sim p_{data}(x)$.

As will be discussed in more detail below, the generators $G_1$ and $G_2$ have the same network structure. From input noise $p_{noise}(z)$, $G_1$ and $G_2$ generate the input network parameters $G_1(z)$ and $G_2(z)$ to be used and evaluated by $E_1$ and $E_2$, respectively. $E_1$ and $E_2$ have the structure of the neural network whose parameters are being optimized or refined. They calculate the classification accuracy on the training data $x$. $E_i(x,G_i(z)), i\epsilon\{1,2\}$ represents the classification accuracy obtained by the evaluator $E_i$ when the parameters $G_i(z)$ are used. The generator resulting in higher accuracy is marked as more accurate generator $G_a$, and the other generator is marked as $G_b$, where $a\in {1,2}, b=!a$.

We define the discriminator $D$ as a network, which is used for binary classification between better and worse generator. $G_1(z)$ and $G_2(z)$ are fed into the discriminator $D$, and the ground truth label about which is the better generator comes from the evaluators. The discriminator $D$ provides the gradients to train the worse performing generator.
\vspace{-0.5cm}
\subsubsection{Generative part} 
\begin{figure}[hbt!]
\centering
 \includegraphics[width=.48\textwidth]{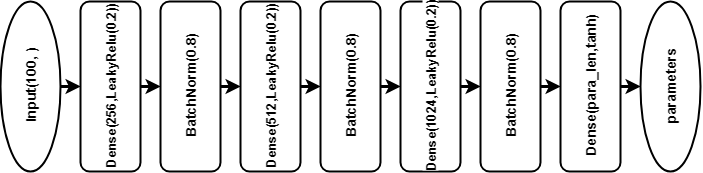}
  \caption{Generator network}
  \label{fig:generator}
\end{figure}

\vspace{-0.1cm}
The two generators $G_1$ and $G_2$ have the same neural network structure shown in Fig.~\ref{fig:generator}. Their input is a Gaussian noise vector $z$, and their outputs are $G_1(z)$ and $G_2(z)$. As seen in Fig.~\ref{fig:generator}, generators are composed of fully connected layers with leaky relu activations. At the output layer, $tanh$ is employed so that $G_i^j(z) \in (-1,1)$, where $j \in \{1,2,...,length(G_i(z))\}$ and $i \in \{1,2\}$. Then, the range of $G_i(z)$ is changed from $(-1,1)$ to $(pm_{min}^j,pm_{max}^j)$ by using
\begin{equation} \label{eqn:range}
G_i(z)'=[G_i(z)\times \frac{pm_{max}-pm_{min}}{2} + \frac{pm_{max}+pm_{min}}{2}].
\end{equation}
In (\ref{eqn:range}), $pm_{max}$ and $pm_{min}$ are preset maxima and minima values, which are defined empirically based on values that a certain parameter can take, so that the value of the refined parameters can only change between $pm_{max}$ and $pm_{min}$. For instance, in the case of the number of neurons for a fully connected layer,  $pm_{max}$ and $pm_{min}$ are 4000 and 10, respectively. The re-scaled values $G_1(z)'$ and $G_2(z)'$ are then used as the parameters of evaluator networks. The length of $G_i(z)$ is determined by the number of network parameters that are refined, and is set at the generator network's last fully connected layer.

Generators are trained/improved by the discriminator, which is a binary classifier used to differentiate the results from generator outputs $G_1(z)$ and $G_2(z)$. Labels ``a" and ``b" represent the generators with higher accuracy and lower accuracy results, respectively. The generator, which has the worse performance and is labeled by ``b", is trained by stochastic gradient descent (SGD) from the discriminator to minimize $log(1-D(G_b(z))$ by using
\begin{equation}
\bigtriangledown_{G_b} \frac{1}{m}\displaystyle\sum_{j=1}^{m} log(1-D(G_b(z^{(j)}))),
\end{equation}
where $m$ is the number of epochs.

When $G_a'(z)$ becomes equal to $G_b'(z)$ for two consecutive iterations, the weights of $G_b$ will be re-initialized to default random values. The purpose of this step is to prevent the optimization stopping at a local maxima and also prevent the vanishing tanh gradient problem.
\vspace{-0.2cm}
\subsubsection{Evaluation part} \vspace{-0.1cm}
As mentioned above, one of the strengths of the proposed approach is that it can be used to refine/optimize parameters of different deep neural network structures. In other words, the evaluator networks have the same structure as the neural network whose parameters are being optimized or refined.

Evaluator networks are built by using the parameters $G_1(z)'$ and $G_2(z)'$ provided by the generators. The training data $x \sim p_{data}(x)$ is used to evaluate these network models. We employ an early stopping criteria. More specifically, if no improvement is observed in $c$ epoches, the training is stopped.

We then obtain the accuracies $acc_i=\mathbb{E}_{x\sim p_{data}(x)}E_i(x)$, $i=\{1,2\}$. Let $a$ be the value of $i$ resulting in higher accuracy, and $b=!a$. Then ``a" is used as the ground truth label for the discriminator, which marks the generator with better parameters, and trains the worse generator $G_b$.
\vspace{-0.2cm}
\subsubsection{Discriminator}
\begin{figure}[hbt!]
\centering
  \includegraphics[width=.4\textwidth]{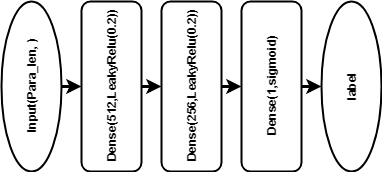}
  \caption{Discriminator network}
  \label{fig:discriminator}
\end{figure}

\vspace{-0.1cm}
We define the discriminator $D$ as a network (seen in Fig.~\ref{fig:discriminator}, whose output is a scalar softmax output, which is used for binary classification between better generator and worse generator. $G_1(z)$ and $G_2(z)$ are fed into the discriminator $D$, and the ground truth label about which is better generator comes from the evaluators. Let $D(G(z))$ represent the probability that $G(z)$ came from the more accurate generator $G_a$ rather than $G_b$. We train $D$ to maximize the probability of assigning the correct label to the outputs $G_1(z)$ and $G_2(z)$ of both generators. Moreover, we simultaneously train the worse generator $G_b$ to minimize $log(1-D(G_b(z))$. The whole process can be expressed by:
\begin{equation} \label{eqn:train_generator}
min_{G_a}max_D \mathbb{E}_{z\sim p_z(z)}(log(D(G_a(z)))+log(1-D(G_b(z)))),
\end{equation}
where, $a=argmax_{i=\{1,2\}} (\mathbb{E}_{x\sim p_{data}(x)}E_i(x))$, $b=!a$.

The pseudo code for the proposed Bi-GAN is provided in Algorithm 1.
\begin{algorithm}[htb]
\caption{Bi-GAN Algorithm.}
\While{in the iterations}{
Generate $m\times2$ noise samples $\{Z_1^{(1)},Z_1^{(2)},...,Z_1^{(m)}\}$ and $\{Z_2^{(1)},Z_2^{(2)},...,Z_2^{(m)}\}$ from Gaussian white noise

\While{j in range(m)}{
Build evaluators $E_1^{(j)}$ and $E_2^{(j)}$ based on parameters from $G_1(Z_1^{(j)})$ and $G_2(Z_2^{(j)})$

Calculate $acc_1^j$ and $acc_2^j$ from $\mathbb{E}_{x\sim p_{data}(x)}E_i(x)$

End if no acc. impr. after $c$ epoches
}
Calculate mean value $acc_i=(1/m)\sum_{j=1}^m acc_i^j$, $i=\{1,2\}$

Find $G_a$ as $G_{argmax(acc_1,acc_2)}$ and $G_b$ as the other one.

Update Discriminator by SGD:
$\bigtriangledown_D \frac{1}{m}\displaystyle\sum_{j=1}^{m} (log(D(G_a(z^{(j)})))+log(1-D(G_b(z^{(j)}))))$

Update Generator $G_b$ by SGD:
$\bigtriangledown_{G_b} \frac{1}{m}\displaystyle\sum_{j=1}^{m} log(1-D(G_b(z^{(j)})))$
}
\label{alg:algorithm_GAN}
\end{algorithm}

\subsection{Genetic Algorithm} \label{ssec:GA}
As mentioned above, we use a GA to make discrete decisions from a set of choices shown in Table~\ref{table:GA parameters}. Within each GA evolution, our proposed Bi-GAN is used to set/optimize the values of the number of filters for the convolutional layers and the number of neurons for the dense layers. Then, the network models are trained and evaluated to obtain their accuracy scores. Based on the accuracy scores, the GA is applied.
\vspace{-0.2cm}
\subsubsection{Initial Population} \vspace{-0.1cm}
The first generation of the networks, $N^1$, is generated randomly such that $N^1 = \{N_1, N_2, ..., N_{n_m}\}$, where $n_m$ is the number of models. This is done by choosing the values of convolutional and dense parameters, in (\ref{eqn:GA_conv_prm}) and (\ref{eqn:GA_dense_prm}), randomly, from the possible choices.
\vspace{-0.2cm}
\subsubsection{Bi-GAN optimization}\vspace{-0.1cm}
Our proposed Bi-GAN is used, as described in Sec.~\ref{ssec:Bigan}, to update the number of neurons for the fully connected layers, and the number of filters for the convolutional layers, of the $n_m$ network models.
\vspace{-0.2cm}
\subsubsection{Evaluation} \vspace{-0.1cm}
After the number of neurons for the fully connected layers, and the number of filters for convolutional layers are determined by Bi-GAN, each generated network model $N_i$ ($i \in \{1,...,n_m\})$ will be evaluated by the fitness function $fitness = f(N_i)$, which is a measure of the accuracy of each model. Models with better performance will have higher values. Thus, $E = \{E_1, E_2, ..., E_{n_m}\}$, will hold the fitness scores $E_i = f(N_i)$, where $i \in \{1, 2, ..., n_m\}$.
\vspace{-0.2cm}
\subsubsection{Selection}\vspace{-0.1cm}
In the selection part, $t$-many top ranked models are selected from the $sorted(E)$ and $r$-many models are selected randomly from the rest of the network models. Then, $d$-many models are dropped in order to prevent over-fitting and getting stuck at a local optimum. The remaining selected models are the parent models ($P$), which will be used to create new models for the next generation.
\vspace{-0.2cm}
\subsubsection{Crossover and Mutation} \vspace{-0.1cm}
Crossover is applied to generate $n_m$-many child network models from the parents. The choice of parents is performed as follows: Instead of always choosing two parents randomly from the parent pool, we associate a counter $C_P$ with each parent $P$, and initialize it to zero. This counter is incremented by one each time a parent is used for crossover. First, two parents are selected randomly from the $t+r-d$ many parents. A new `child' network is generated from the parents via crossover, and the counters of the parents are incremented by one. Then, two parents, whose counter is still zero, are selected randomly from the parent pool. Another network is generated from them via crossover, and the counters of the parents are incremented. If there is only one network model left with counter equal to zero, and the number of children is still less than $n_m$, then this model is chosen as one of the parents, and the other parent is chosen randomly from the rest of the models who have a counter value of one. If there are no more parents left with counter equal to zero, and the number of children is still less than $n_m$, then two parents, whose counter is one, are picked randomly, and their counter is incremented to two after crossover. This process is repeated until the number of children models reaches $n_m$.

\begin{figure*}
\centering
\includegraphics[width=.6\textwidth]{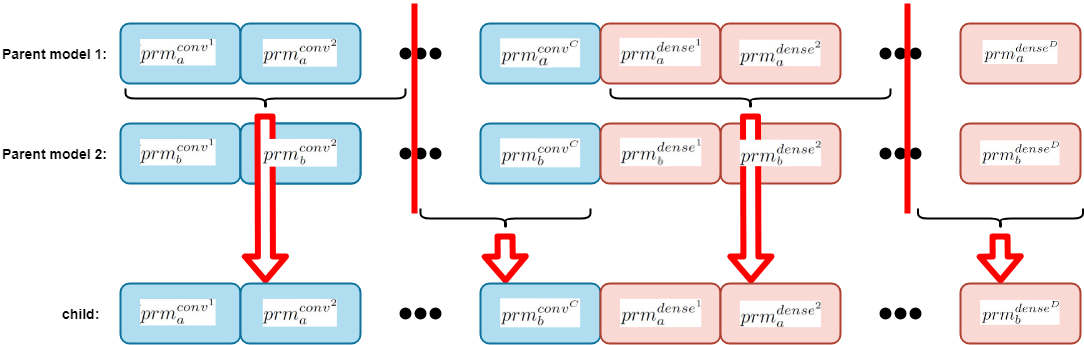}
\caption{Crossover.}
\label{fig:Crossover}
\end{figure*}

The crossover between parent models $a$ and $b$ is performed as illustrated in Fig.~\ref{fig:Crossover}. First two integers (ID$_1$ and ID$_2$) are picked randomly between $1 \text{ and } C$ and $1 \text{ and } D$, respectively. Then, the parameters of the child network is set so that
\begin{equation}
\begin{split}
p_{child}^{GA}=&[prm_a^{conv^1}, prm_a^{conv^2}, ... , prm_a^{conv^{ID_1}}, \\
&prm_b^{conv^{ID_1+1}},..., prm_b^{conv^C}, \\
&prm_a^{dense^1}, prm_a^{dense^2}, ... , prm_a^{dense^{ID_2}}, \\
&prm_b^{dense^{ID_2+1}},..., prm_b^{dense^D}].
\end{split}
\label{eqn:GA_child}
\end{equation}

After all the $n_m$-many child networks are obtained via crossover, 20\% of the population is chosen randomly to perform mutation.  As seen in (\ref{eqn:GA_conv_prm}) and (\ref{eqn:GA_dense_prm}), there are five different convolutional layer parameters, and four different dense layer parameters. Thus, there are $5*C + 4*D$-many possible parameters that can be mutated. An integer is picked randomly between $1$ and $5*C + 4*D$, and the corresponding parameter type is chosen randomly from the possible choices in Table \ref{table:GA parameters}. For instance, if the random number corresponds to the filter size parameter, then its value is chosen randomly from $\{3,5\}$.

Then, the entire process is repeated by using this new population, updating the number of neurons and the number of filters for each network model in the population by using our propose Bi-GAN, and so on. The pseudo code for the entire process is provided in Algorithm \ref{alg:algorithm_total}.

\begin{algorithm}[htb]
\caption{GA-BiGAN algorithm.}

Randomly initialize $n_m$ models for population $N^1$.

\While{$i^{th}$ iteration}{
Update hyper parameters by Bi-GAN (Alg.~\ref{alg:algorithm_GAN}).

Train and evaluate $N^i_1, N^i_2, ..., N^i_{n_m}$ by fitness function $f(N^i_j)$ and obtain scores $E$.

Select $t$ top scored networks $N_{top} = N^i(argmax(E))$

Randomly choose $r$ networks $N_{rand}$ from the rest of population $N^i$

Merge $N_{top}$ and $N_{rand}$ and then drop $d$ networks ($N_{drop}$)

Form $N_{parent} = {(N_{top} \bigcup N_{rand}) - N_{drop} }$

Choose parents from $N_{parent}$ for crossover and generate $n_m$ new networks and add them to $N^{i+1}$

Choose $20\%$ of the networks in $N^{i+1}$, and perform mutation on them.
}

\label{alg:algorithm_total}
\end{algorithm}
\vspace{-0.2cm}
\section{Experimental Results} \label{sec:exp}
\begin{figure}[hbt!]
\centering
\begin{minipage}{.12\textwidth}
\centering
\includegraphics[width=1\linewidth]{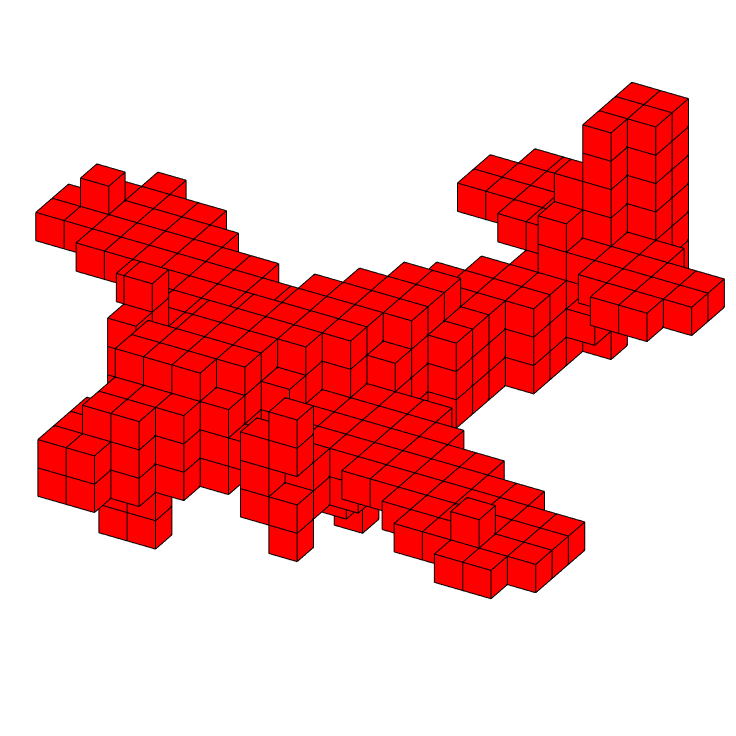}
\end{minipage}%
\begin{minipage}{.12\textwidth}
\centering
\includegraphics[width=1\linewidth]{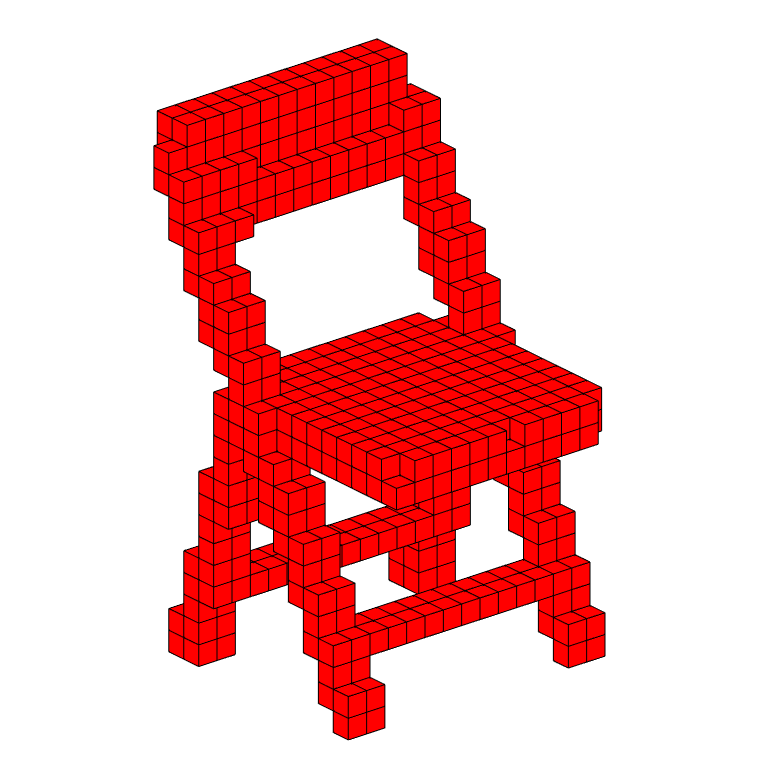}
\end{minipage}
\begin{minipage}{.12\textwidth}
\centering
\includegraphics[width=1\linewidth]{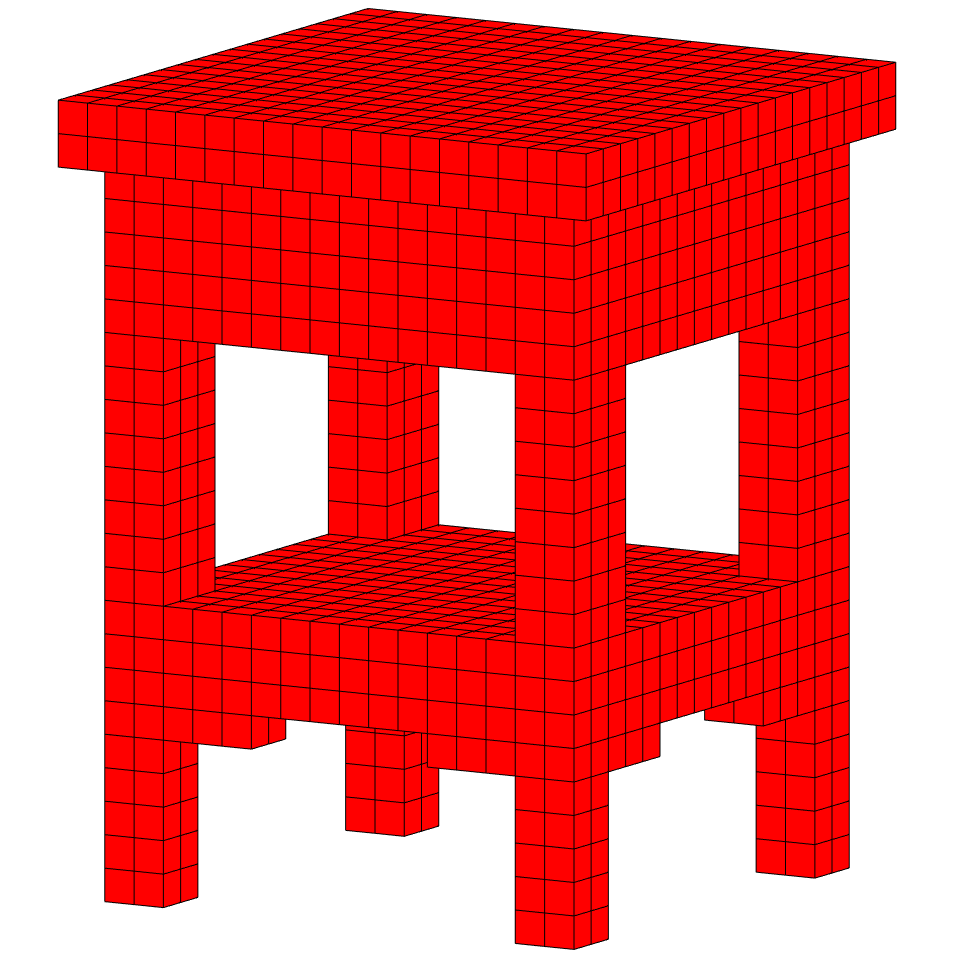}
\end{minipage}
\begin{minipage}{.12\textwidth}
\centering
\includegraphics[width=1\linewidth]{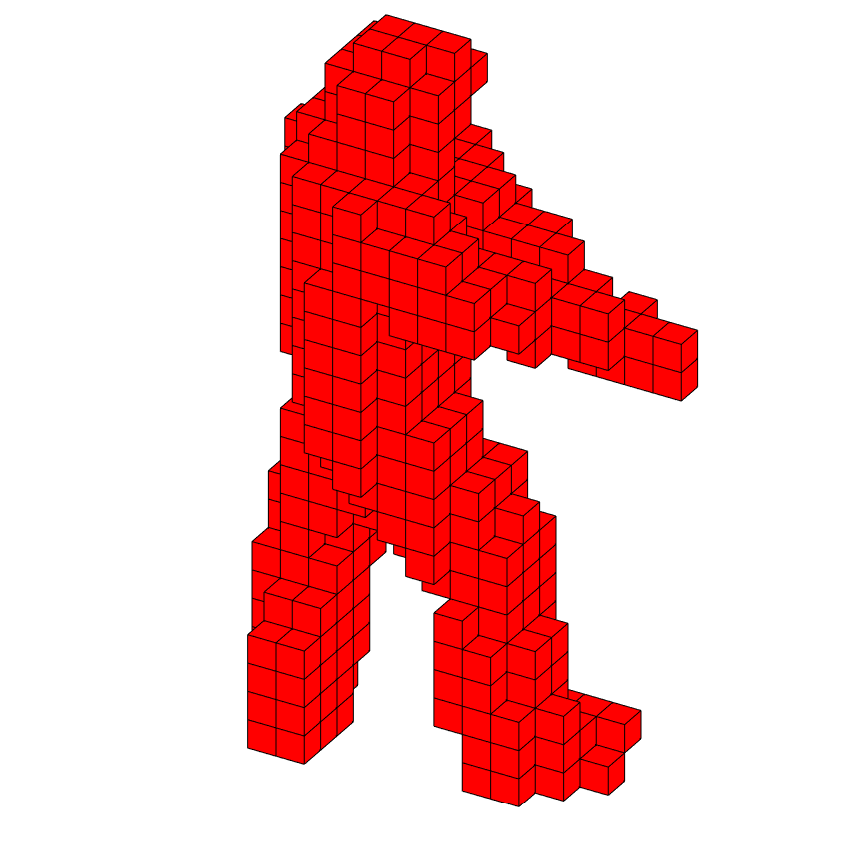}
\end{minipage}
\begin{minipage}{.12\textwidth}
\centering
\includegraphics[width=1\linewidth]{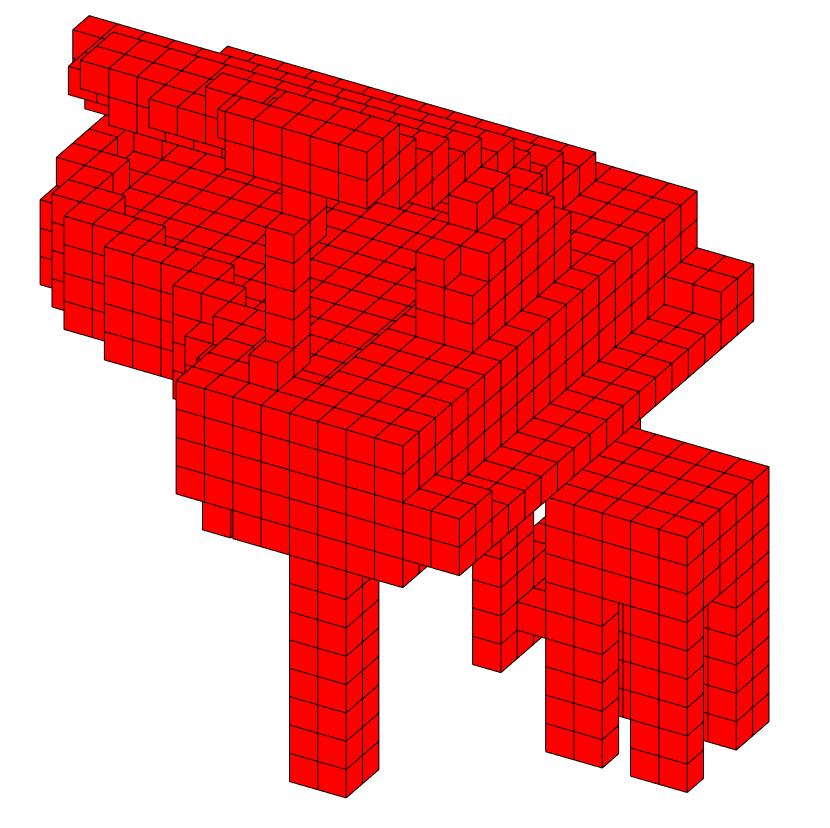}
\end{minipage}
\caption{Sample voxelized objects from ModelNet40 dataset.}
\label{fig:exp_modelnet}
\vspace{-0.1cm}
\end{figure}

Without loss of generality, we have applied the proposed approach on a 3D convolutional neural network by using the voxelized version of ModelNet40 dataset, which contains 3D CAD models of 40 object classes. Wu et al.~\cite{wu20153d} voxelized each object from the ModelNet at 12 different orientations (around gravity axis) for data augmentation. Voxel grids are  $30 \times30 \times30$, and every object is fitted into this range. We used this pre-voxelized version of the dataset in our experiments. The dataset contains 40 subfolders for different objects. Each of these 40 subfolders contains 2 subfolders for training and testing. Train:test ratio differs for each object, but overall train:test ratio is around 3:1. Some example voxelized objects from the ModelNet40 dataset are shown in Fig.~\ref{fig:exp_modelnet}.

We have compared our proposed approach with two other approaches, which are based only on GAs and referred to as small-set GA, and large-set GA. The small-set GA is a basic genetic algorithm with limited number of choices. Large-set GA is given a larger set of choices for the number of neurons, and the number of filters. As for the activation function, batch normalization, dropout and max pooling decisions, the number of convolutional and dense layers, and the kernel size, the parameter choices are the same as in Table \ref{table:GA parameters} for both small- and large-set GA. The difference between small- and large-set GA is the parameter choices for the number of neurons and the number of filters. For the small-set GA these choices are as follows:
\begin{table}[h!]
{\small{
\vspace{0.1cm}
\begin{tabular}{ll}
Num. of neurons: & \{16,32,64,128,256,512,1024,2048,4096\}  \\
Num. of kernels: & \{1,4,16,64,256\}
\end{tabular}
\label{table:GA parameters}
}}
\vspace{-0.2cm}
\end{table}

For the large-set GA, the number of neurons can be any integer between $16$ and $4096$, and the number of filters can be any integer between $1$ and $256$. In our proposed approach, the number of neurons and the number of filters are determined by our proposed Bi-GAN method from this continuous range.

\begin{figure}[h!]
\vspace{-0.2cm}
  \centering
   \centerline{\includegraphics[width=.38\textwidth]{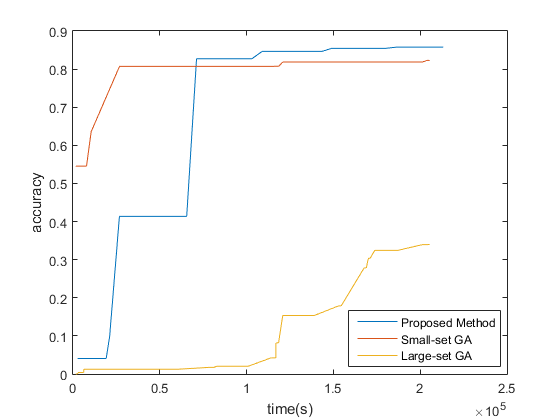}}
   \vspace{-0.1cm}
\caption{Accuracy of different network refinement approaches over time.}
\label{fig:time_acc}
\vspace{-0.2cm}
\end{figure}

These three approaches were run by using the same data, for the same amount of time to compare their performances. The parameters used in Algorithm 1 and 2 are as follows: For the Bi-GAN part, $m=100$, and $c=5$. For the GA part, the parameters used are $n_m=25$, $t=4$, $r=2$ and $d=1$. The results are summarized in Table \ref{table:55hour}. Same population size was used for all the GAs. As can be seen, our proposed approach provides the highest accuracy, and performs better compared to only GA-based approaches. Figure \ref{fig:time_acc} shows the accuracy of each method over time. The proposed method determines the number of neurons and the number of filters without requiring a discrete set of choices.
\begin{table}[h!]
\vspace{-0.2cm}
\centering
\caption{Accuracy values obtained with different networks}
\begin{tabular}{|c|c|}
\hline
                & Accuracy \\ \hline
ShapeNet model    & 0.8417   \\
Small-set GA      & 0.8294   \\
Large-set GA      & 0.3641   \\
Proposed Method   & \textbf{0.8520}   \\ \hline
\end{tabular}
\label{table:55hour}
\vspace{-0.2cm}
\end{table}

Another important point is the performance comparison with respect to Shapenets~\cite{wu20153d}, which is a hand-crafted network with four convolutional layers, and two dense layers. Table \ref{table:55hour} also includes the accuracy obtained when the Shapenet~\cite{wu20153d} model is used. As can be seen, by autonomously refining the network parameters, our proposed approach provides higher accuracy than the manually designed model by using a shallower network. The parameter values that each of the three approaches (proposed method, small-set GA and large-set GA) ends up choosing/using are provided in Table \ref{table:55hourparam} (the first entry is the filter size). As can be seen, the proposed method achieves this higher accuracy of 85.2\% by using only three convolutional layers as opposed to the four-convolutional-layer Shapenet model.
\begin{table*}[htb!]
\begin{center}
\begin{tabular}{cllcccc}
\hline
Iterations & \multicolumn{2}{c}{5-pops} & \multicolumn{2}{c}{10-pops} & \multicolumn{2}{c}{20-pops} \\ \hline
           & Small-set GA      & Proposed      & Small-set GA      & Proposed       & Small-set GA      & Proposed       \\
10         & 0.5813        & 0.6077     & 0.5945        & 0.6563      & 0.6409        & 0.7945      \\
20         & 0.6969        & 0.7601     & 0.6183        & 0.7798      & 0.7689        & 0.7984      \\
50         & 0.7626        & 0.8124     & 0.8069        & 0.8197      & 0.8113        & 0.8429      \\ \hline
\end{tabular}
\end{center}
\caption{Comparison of the proposed method with the Small-set GA for different population sizes.}
\label{table:GA_comp}
\end{table*}

\begin{table}[hbt!]
\vspace{-0.2cm}
\caption{}
\small{
\begin{tabular}{|l|l|}
\hline
 Approach             & Final Parameter Values  \\ \hline
Sm-set GA       &[C1: 5,`Relu', NO Batch norm,Maxpool,64], \\
                &[C2: 5,`Relu',No Batch,Max pool,64], \\
                &[C3: 3,`Relu',0,1,256],  \\
                &[D1: Relu, No Batch, Dropout, 256],  \\
                &[D2: Leaky relu, no batch, no dropout, 64]  \\ \hline
Lrg-set GA    & [C1: 5,'Leaky Relu',No Batch n., max pool,30], \\
                &[C2: 3,`Relu',No Batch, Max pool, 31], \\
                &[C3: 5,`Relu',No batch, No max pool, 9],  \\
                &[D1: Leaky Relu, Batch norm, Dropout, 31]  \\ \hline
Prop.Meth. & [C1: 5,'Relu',No Batch n,max pool,80], \\
                &[C2: 3,`Relu',No Batch,Max pool, 105], \\
                &[C3: 3,`Leaky Relu',No batch n.,Max pool,202],  \\
                &[D1: relu, no batch, no dropout, 601],  \\
                &[D2:leaky relu, no batch, dropout, 240]  \\ \hline
\end{tabular}
}
\label{table:55hourparam}
\vspace{-0.2cm}
\end{table}

\begin{figure}[ht!]
  \centering
   \centerline{\includegraphics[width=.4\textwidth]{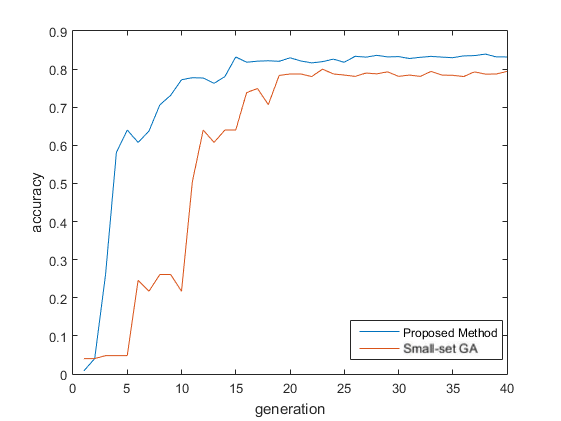}}
   \vspace{-0.11cm}
\caption{Accuracy with every evolution of the proposed method and the Small-set GA.}
\label{fig:training acc_1_GAGAN_modelnet}
\vspace{-0.4cm}
\end{figure}

\begin{figure}[ht!]
  \centering
   \centerline{\includegraphics[width=.4\textwidth]{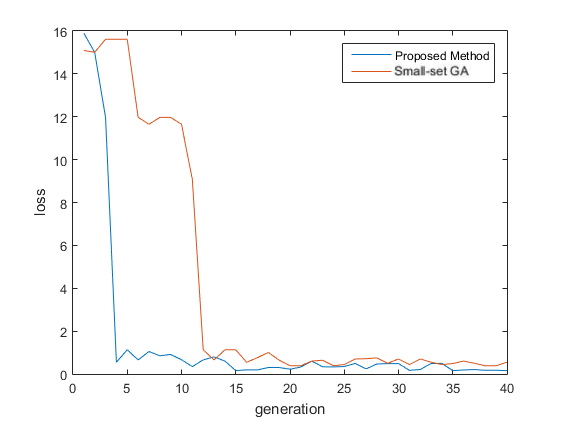}}
   \vspace{-0.1cm}
\caption{Loss with every evolution of the proposed method and the Small-set GA.}
\label{fig:training loss_1_GAGAN_modelnet}
\vspace{-0.4cm}
\end{figure}

Since small-set GA performs better than the large-set GA, in the remainder of the experiments, we compared our proposed method with the small-set GA. We have tried three more population sizes while keeping the other parameter choices same as before. The results are summarized in Table \ref{table:GA_comp}. As can be seen, the proposed method provides the higher accuracy rates for all different population sizes. Figures \ref{fig:training acc_1_GAGAN_modelnet} and \ref{fig:training loss_1_GAGAN_modelnet} show the change in accuracy and loss of these approaches, respectively, with each evolution (when population size is 20). As can be seen, the proposed method performs better during its evolutions.

We also performed one more experiment, where we let the maximum number of convolutional layers to be 5. In this experiment, the proposed method achieved an accuracy of 86.62\%.



\section{Conclusion}\label{sec:concl}
In this paper, we have presented a novel and systematic method that autonomously and simultaneously optimizes multiple parameters of any given deep neural network by using a genetic algorithm (GA) aided by a novel Bi-Generative Adversarial Network (GAN) with two generators, which is referred to as Bi-GAN. The proposed Bi-GAN allows the autonomous exploitation and choice of the number of neurons, for the fully-connected layers, and number of filters for the convolutional layers, from a large range of values. Our proposed approach can be used to autonomously refine the number of convolutional layers and dense layers, number and size of kernels, and the number of neurons; choose the type of the activation function; and decide whether to use dropout and batch normalization or not, to improve the accuracy of different deep neural network architectures. Without loss of generality, the proposed method has been tested with the ModelNet database, and compared with the 3D Shapenets and two GA-only methods. The results show that the presented approach can simultaneously and successfully optimize multiple neural network parameters, and achieve increased accuracy even with shallower networks.

\newpage
\bibliographystyle{IEEEbib}
\pagebreak
\bibliography{egbib}
\end{document}